\documentclass[a4,a4wide]{article}
\usepackage{aaai}
\usepackage{times}
\usepackage{helvet}
\usepackage{courier}
\usepackage{latexsym} 
\usepackage{amssymb}

\usepackage{xspace}
\usepackage{subfigure}

\usepackage{url}

\usepackage{epsfig}
\usepackage{graphicx}

\newcommand{\clasp}{{\sc clasp}\xspace}
\newcommand{\claspD}{{\sc claspD}\xspace}
\newcommand{\claspf}{{\sc claspfolio}\xspace}

\newcommand{\cmodels}{{\sc cmodels}\xspace}
\newcommand{\dlv}{{\sc DLV}\xspace}
\newcommand{\idp}{{\sc idp}\xspace}

\newcommand{\gringo}{{\sc GrinGo}\xspace}
\newcommand{\measp}{{\sc me-asp}\xspace}
\newcommand{\measpGR}{{\sc me-asp}$^{gr}$\xspace}
\newcommand{\measpOLD}{{\sc me-asp}$^{old}$\xspace}
\newcommand{\measpNEW}{{\sc me-asp}$^{new}$\xspace}

\newcommand{\sota}{\textsc{sota}\xspace}

\newcommand{\Pol}{{\em P}\xspace}
\newcommand{\NP}{{\em NP}\xspace}
\newcommand{\BNP}{{\em Beyond NP}\xspace}

\newcommand{\myparagraph}[1]{\vspace{0.06cm}\noindent{\bf #1}\xspace}

\newcommand{\citeASP}{\cite{bara-2002,eite-etal-97f,gelf-lifs-88,gelf-lifs-91,mare-trus-98sm,niem-98}\xspace}
\newcommand{\citeASPComp}{\cite{cali-etal-2011-syscomp}\xspace}

\begin{document}
\nocopyright


\title{The Multi-engine ASP Solver ME-ASP: Progress Report\thanks{This paper will appear in the Proceedings of the 15th International Workshop on Non-Monotonic Reasoning (NMR 2014)}}

\author{Marco Maratea \\
DIBRIS, \\
Univ. degli Studi di Genova, \\
Viale F. Causa 15, 16145 Genova, Italy\\
\texttt{marco@dist.unige.it} \\
\And
Luca Pulina \\
POLCOMING, \\
Univ. degli Studi di Sassari, \\
Viale Mancini 5, 07100 Sassari, Italy\\
\texttt{lpulina@uniss.it} \\
\And
Francesco Ricca \\
Dip. di Matematica ed Informatica, \\
Univ. della Calabria, \\
Via P. Bucci, 87030 Rende, Italy \\
\texttt{ricca@mat.unical.it}
}

\newcommand{\TODO}[1]{{\bf\large TODO:} {\em #1 }}




\maketitle

\begin{abstract}
\measp is a multi-engine solver for ground ASP programs. It exploits algorithm selection techniques based on classification
to select one among a set of out-of-the-box heterogeneous ASP solvers used as black-box engines.
In this paper we report on $(i)$ a new optimized implementation of \measp; and
$(ii)$ an attempt of applying algorithm selection to non-ground programs.  
An experimental analysis reported in the paper
shows that $(i)$ the new implementation of \measp is substantially
faster than the previous version; and $(ii)$ the multi-engine recipe 
can be applied to the evaluation of non-ground programs with some
benefits.
\end{abstract}



\section{Introduction}\label{sec:intro}
Answer Set Programming~\citeASP (ASP) is a
declarative language based on logic programming and non-monotonic
reasoning.  The applications of ASP belong to several areas, 
e.g., ASP was used for solving a variety of hard combinatorial problems
(see e.g.,~\citeASPComp and ~\cite{asparagus}).

Nowadays, several efficient ASP systems are available%
~\cite{gebs-etal-2007-ijcai,janh-etal-2009,leon-etal-2002-dlv,lier-2005-lpnmr,mari-etal-2008,simo-etal-2002}.
It is well-established that, for solving empirically hard problems, there is
rarely a best algorithm/heuristic, while it is often the case that
different algorithms perform well on different problems/instances.  
It can be easily verified (e.g., by analyzing the results of the ASP
competition series) that this is the case also for ASP implementations.
In order to take advantage of this fact, one should be able to select
automatically the ``best'' solver on the basis of the characteristics
(called \textsl{features}) of the instance in input, i.e., one has to
consider to solve an \textsl{algorithm selection problem}~\cite{rice-76}.

Inspired by the successful attempts%
~\cite{gome-selm-2001,omah-etal-08,pul09a,xu-etal-2008} 
done in the neighbor fields of SAT, QSAT and CSP, the application of
algorithm selection techniques to ASP solving was ignited by the
release of the portfolio solver \claspf~\cite{gebs-etal-2011-claspfolio}.
This solver imports into ASP the {\sc satzilla}~\cite{xu-etal-2008}
approach.  Indeed, \claspf employs inductive techniques 
based on {\em regression} to choose the ``best'' configuration/heuristic 
of the solver {\clasp}.  
The complete picture of inductive approaches applied to ASP solving
includes also techniques for learning heuristics
orders~\cite{bald-11}, solutions to combine portfolio and automatic
algorithm configuration approaches~\cite{silv-etal-2012-iclp},
automatic selection of a scheduling of ASP
solvers~\cite{hoos-etal-2012-iclp} (in this case {\clasp}
configurations), and the multi-engine approach.  
The aim of a multi-engine solver~\cite{pul09a} is to select the ``best''
solver among a set of efficient ones used as {\em black-box engines}.
The multi-engine ASP solver \measp was proposed in%
~\cite{mara-etal-2012-jelia}, and ports to ASP 
an approach applied before to QBF~\cite{pul09a}.  

\measp exploits inductive techniques based on {\em classification} to
choose, on a per instance basis, an engine among a selection of
black-box heterogeneous ASP solvers.
%
The first implementation of \measp,
despite not being highly optimized, already reached good performance.
Indeed, \measp can combine the strengths of its component engines, and
thus it performs well on a broad set of benchmarks including 14
domains and 1462 ground instances (detailed results are reported in~\cite{maratea2014multi}).

In this paper we report on $(i)$ a new optimized implementation of \measp; and on
$(ii)$ a first attempt of applying algorithm selection to the entire process of 
computing answer sets of non-ground programs.

As a matter of fact, the ASP solutions available at the state of the art employing
machine-learning techniques are devised to solve ground (or
propositional) programs, and -- to the best of our knowledge -- no
solution has been proposed that is able to cope directly with
non-ground programs.  Note that ASP programmers almost always write
non-ground programs, which have to be first instantiated by a
grounder.  It is well-known that such instantiation phase can
influence significantly the performance of the entire solving process.
At the time of this writing, there are two prominent alternative
implementations that are able to instantiate ASP programs:
\dlv~\cite{leon-etal-2002-dlv} and \gringo~\cite{gebs-etal-2007-gringo}.
Once the peculiarities of the instantiation process are properly taken into account, both
implementations can be combined in a multi-engine grounder by applying
also to this phase an algorithm selection recipe, building on~\cite{maratea2013automated}.
The entire process of evaluation of a non-ground ASP program can be, thus, obtained 
by applying algorithm selection to the instantiation phase, selecting either \dlv or \gringo; and,
then, in a subsequent step, evaluating the propositional program obtained in the first step with a multi-engine solver.

An experimental analysis reported in the paper shows that $(i)$ the
new implementation of \measp is substantially faster than the previous
version; and $(ii)$ the straight application of the multi-engine
recipe to the instantiation phase is already beneficial.  At the same
time, it remains space for future work, and in particular for devising
more specialized techniques to exploit the full potential of the
approach.


\section{A Multi-Engine ASP system}\label{sec:prelim}

We next overview the components of the multi-engine approach, and we
report on the way we have instantiated it to cope with instantiation and solving,
thus obtaining a complete multi-engine system for computing answer sets of
non-ground ASP programs.

\myparagraph{General Approach.} The design of a multi-engine solver based on classification is composed
of three main ingredients: $(i)$ a set of features that are
significant for classifying the instances; $(ii)$ a selection of
solvers that are representative of the state of the art and complementary; and $(iii)$ a
choice of effective classification algorithms.  Each instance in a
fairly-designed \textsl{training set} of
instances is analyzed by considering both
the features and the performance of each solvers. 
An inductive model is computed by the classification algorithm during
this phase.  Then, each instance in a \textsl{test set} is processed
by first extracting its features, and the solver is selected starting
from these features using the
learned model. Note that, this schema does not make any assumption
(other than the basic one of supporting a common input) on the
engines.


\myparagraph{The \measp solver.}
In~\cite{mara-etal-2012-jelia,maratea2014multi} we described the
choices we have made to develop the \measp solver. In particular, we
have singled out a set of syntactic features that are both significant
for classifying the instances and cheap-to-compute (so that the
classifier can be fast and accurate).  In detail, we considered:
the number of rules and number of atoms, the ratio of horn, unary,
binary and ternary rules, as well as some ASP peculiar features, such
as the number of true and disjunctive facts, and the fraction of
normal rules and constraints. The number of resulting features, together with some of their combinations, amounts to 52.  In
order to select the engines we ran preliminary
experiments~\cite{maratea2014multi} to collect a pool of solvers that
is representative of the state-of-the-art solver ({\sc sota}), i.e.,
considering a problem instance, the oracle that always fares the best
among the solvers that entered the system track of the 3rd ASP
Competition~\cite{cali-etal-2011-syscomp}, plus {\dlv}.  The pool of
engines collected in {\measp} is composed of 5 solvers, namely \clasp,
\claspD, \cmodels, \dlv, and \idp, as submitted to the 3rd ASP
Competition.  We experimented with several
classification algorithms~\cite{maratea2014multi}, and proved
empirically that \measp can perform better than its engines with any
choice.  Nonetheless, we selected the k-nearest neighbor (kNN)
classifier for our new implementation: it was already used 
in {\measp}~\cite{mara-etal-2012-jelia}, with good
performance, and it was easy to integrate its implementation in the
new version of the system.


\myparagraph{Multi-engine instantiator.}
Concerning the automatic selection of the grounder, we selected: 
number of disjunctive rules,
presence of queries, the total amount of functions and predicates,
number of strongly connected and Head-Cycle Free\cite{bene-dech-94} components,
and stratification property, for a total amount of 11 features.
These features are able to discriminate the class of the problem, and
are also pragmatically cheap-to-compute.  Indeed, given the high
expressivity of the language, non-ground ASP programs (which are
usually written by programmers) contain only a few rules.  Concerning
the grounders, given that there are only two alternative solutions,
namely \dlv and \gringo, we considered both for our implementation.

Concerning the classification method, we used an implementation of the
PART decision list generator~\cite{fran-witt-icml-98}, a classifier
that returns a human readable model based on if-then-else rules. We
used PART because, given the relatively small total amount of features
related to the non-ground instances, it allows us to compare
the generated model with respect to the knowledge of a human expert.

\myparagraph{Multi-Engine System \measpGR.}
Given a (non-ground) ASP program, the evaluation workflow of 
the multi-engine ASP solution called \measpGR\ is the following:
$(i)$ non-ground features extraction, $(ii)$ grounder selection,
$(iii)$ grounding phase, $(iv)$ ground features extraction,
$(v)$ solver selection, and $(vi)$ solving phase on ground program.

\section{Implementation and Experiments}\label{sec:experiments}\label{sec:benchs}

In this section we report the results of two experiments conceived to
assess the performance of the new versions of the \measp system.  The
first experiment has the goal of measuring the performance
improvements obtained by the new optimized implementation of the
\measp solver.  The second experiment assesses \measpGR and reports on
the performance improvements that can be obtained by selecting the
grounder first and then calling the \measp solver.  {\measp} and
\measpGR are available for download at
\url{www.mat.unical.it/ricca/me-asp}.  Concerning the hardware
employed and the execution settings, all the experiments run on a
cluster of Intel Xeon E31245 PCs at 3.30 GHz equipped with 64 bit
Ubuntu 12.04, granting 600 seconds of CPU time and 2GB of memory to
each solver.  The benchmarks used in this paper belong to the suite of
benchmarks, encoded in the ASP-Core 1.0 language, of the 3rd ASP
Competition.  Note that in the 4th ASP
Competition~\cite{alvi-etal-2013-comp} the new language ASP-Core 2.0
has been introduced.  We still rely on the language of the 3rd ASP
Competition given that the total amount of solvers and grounders
supporting the new standard language is very limited with respect to
the number of tools supporting ASP-Core 1.0.

\myparagraph{Assessment of the new implementation of \measp.} 
The original implementation of \measp was obtained by combining a 
general purpose feature extractor (that we have initially developed
for experimenting with a variety of additional features) developed in Java,
with a collection of Perl scripts linking the other components 
of the system, which are based on the {\em rapidminer} library.
This is a general purpose implementation supporting also several
classification algorithms.
Since the CPU time spent for the extraction of features and solver selection
has to be made negligible, we developed an optimized version of \measp.
The goal was to optimize the interaction among system components 
and further improve their efficiency.
To this end, we have re-engineered the feature extractor, enabling it
to read ground instances expressed in the numeric format used by
{\gringo}. Furthermore, we have integrated it with an implementation
of the kNN algorithm
built on top of the
ANN library (\url{www.cs.umd.edu/~mount/ANN}) in the same
binary developed in C++.  This way the new implementation minimizes
the overhead introduced by solver selection.

\begin{figure}[t]

\begin{center}
  \begin{tabular}{c}
    \scalebox{0.525}{\includegraphics{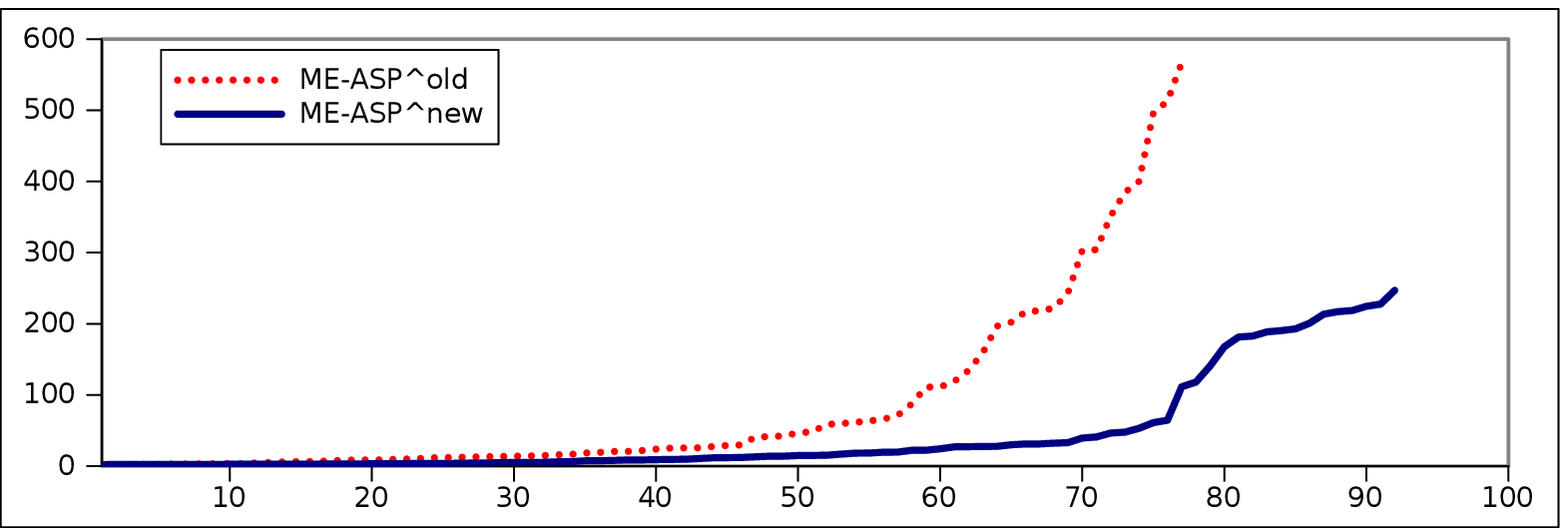}} \\
    \scalebox{0.525}{\includegraphics{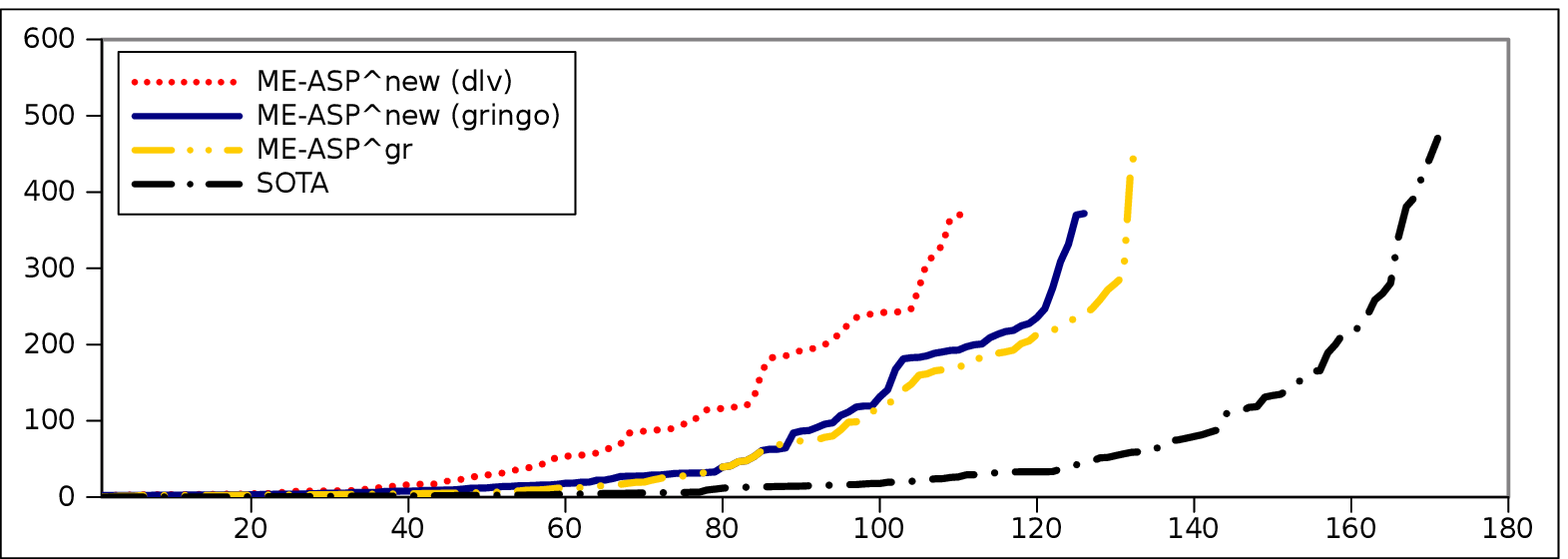}} \\
  \end{tabular}
\end{center}
  \caption{\small{Performance of \measpOLD and \measpNEW on {\NP} and  {\BNP}
    instances evaluated at the 3rd ASP Competition (top); performance
    of \measpGR, its engines and {\sota} on the complete set
    of instances evaluated at the 3rd ASP Competition (bottom). In the
    $x$-axis it is shown the total amount of solved instances, while
    $y$-axis reports the CPU time in seconds.} }
  \label{fig:plot}

\end{figure}

We now present the results of an experiment in which we compare the
old implementation of {\measp}, labelled \measpOLD, with the new one,
labelled \measpNEW.  In this experiment, assessing solving
performance, we used \gringo as grounder for both implementations, and
we considered problems belonging to the {\NP} and {\BNP} classes of
the competition (i.e., the grounder and domains considered by
\measpOLD~\cite{maratea2014multi}). The inductive model used in
\measpNEW\ was the same used in \measpOLD (details are reported
in~\cite{maratea2014multi}).  The plot in Figure~\ref{fig:plot} (top)
depicts the performance of both \measpOLD and \measpNEW (dotted red
and solid blue lines in the plot, respectively). Considering the total
amount of {\NP} and {\BNP} instances evaluated at the 3rd ASP
Competition (140), \measpNEW solved 92 instances (77 {\NP} and 15
{\BNP}) in about 4120 seconds, while \measpOLD solved 77 instances (62
{\NP} and 15 {\BNP}) in about 6498 seconds. We report an improvement
both in the total amount of solved instances (\measpNEW is able to
solve 66\% of the whole set of instances, while \measpNEW stops at
51\%) and in the average CPU time of solved instances (about 45
seconds against 84).

The improvements of \measpNEW are due to its optimized implementation.
Once feature extraction and solver selection are made very efficient,
it is possible to extract features for more instances and the
engines are called in advance w.r.t. what happens in \measpOLD.
  This results in more instances that are processed and solved by \measpNEW
within the timeout.

\myparagraph{Assessment of the complete system.} 
We developed a preliminary implementation of a grounder selector,
which combines a feature extractor for non-ground programs written in
Java, and an implementation of the PART decision list generator, as
mentioned in the previous section. The grounder selector is then
combined with \measpNEW.

We now present the results of an experiment in which we compare
\measpGR with \measpNEW, and the \sota solver.  \measpNEW coupled with
\dlv (resp. \gringo) is denoted by \measpNEW(dlv)
(resp. \measpNEW(gringo)). In this case we considered all the
benchmark problems of the 3rd ASP Competition, including the ones
belonging to the {\Pol} class.  Indeed, in this case we are interested
also in grounders' performance, which is crucial in the {\Pol} class.

The plot in Figure~\ref{fig:plot} (bottom) shows the performance of
the aforementioned solvers. In the plot, we depict the performance of
\measpNEW(dlv) with a red dotted line, \measpNEW(gringo) with a solid
blue line, \measpGR with a double dotted dashed yellow line, and,
finally, with a dotted dashed black line we denote the performance of
the \sota solver.  Looking at the plot, we can see that
\measpNEW(gringo) solves more instances that \measpNEW(dlv) -- 126 and
111, respectively -- while both are outperformed by \measpGR, that is
able to solve 134 instances.  The average CPU time of solved instances
for \measpNEW(dlv), \measpNEW(gringo) and \measpGR is 86.86, 67.93 and
107.82 seconds, respectively.  Looking at the bottom plot in
Figure~\ref{fig:plot}, concerning the performance of the \sota solver,
we report that it is able to solve 173 instances out of a total of 200
instances (evaluated at the 3rd ASP Competition), highlighting room
for further improving this preliminary version of {\measpGR}.  Indeed,
the current classification model predicts \gringo for most of the
{\NP} instances, but having a more detailed look at the results, we
notice that \clasp and \idp with \gringo solve both 72 instances,
while using \dlv they solve 93 and 92 instances, respectively.  A
detailed analysis of the performance of the various ASP solvers with
both grounders can be found in~\cite{maratea2013automated}.

It is also worth mentioning that the output formats of \gringo\ and \dlv differ, thus
there are combinations grounder/solver that require additional conversion steps in our implementation.
Since the new feature extractor is designed to be compliant with the numeric format 
produced by \gringo, if \dlv\ is selected as grounder then the non-ground program is instantiated twice.
Moreover, if \dlv\ is selected as grounder, and it is not selected also as solver, 
the produced propositional program is fed in gringo to be converted in numeric format. 
These additional steps, due to technical issues, result in a
suboptimal implementation of the execution pipeline that could be
further optimized in case both grounders would agree on a common
output format.

\medskip
\noindent{\bf Conclusion.}  
In this paper we presented improvements to
the multi-engine ASP solver {\measp}. Experiments show that $(i)$ the
new implementation of \measp is more efficient, and $(ii)$ the
straight application of the multi-engine recipe to the instantiation
phase is already beneficial. Directions for future research include
exploiting the full potential of the approach by predicting
the pair grounder+solver, and importing policy adaptation
techniques employed in~\cite{maratea2014adapt}.

\medskip
\noindent{\bf Acknowledgments.} This research has been partly
supported by Regione Calabria under project PIA KnowRex POR FESR 2007-
2013 BURC n. 49 s.s. n. 1 16/12/2010, the Italian Ministry of
University and Research under PON project ``Ba2Know S.I.-LAB''
n. PON03PE\_0001, the Autonomous Region of Sardinia (Italy) and the Port Authority of Cagliari (Italy)
under L.R. 7/2007, Tender 16 2011 project ``\textsc{desctop}'', 
CRP-49656.

\bibliographystyle{aaai}
\newcommand{\SortNoOp}[1]{}

\end{document}